# Sequential convolutional network for behavioral pattern extraction in gait recognition

Xinnan Ding[a], Kejun Wang[a,*], Chenhui Wang[b], Tianyi Lan[a], Liangliang Liu[a]

[a] College of Intelligent Systems Science and Engineering, Harbin Engineering University, Harbin 150001, China
[b] Department of Statistics, University of California, Los Angeles, CA 90095-1554, USA

**Abstract**

As a unique and promising biometric, video-based gait recognition has broad applications. The key step of this methodology is to learn the walking pattern of individuals, which, however, often suffers challenges to extract the behavioral feature from a sequence directly. Most existing methods just focus on either the appearance or the motion pattern. To overcome these limitations, we propose a sequential convolutional network (SCN) from a novel perspective, where spatiotemporal features can be learned by a basic convolutional backbone. In SCN, behavioral information extractors (BIE) are constructed to comprehend intermediate feature maps in time series through motion templates where the relationship between frames can be analyzed, thereby distilling the information of the walking pattern. Furthermore, a multi-frame aggregator in SCN performs feature integration on a sequence whose length is uncertain, via a mobile 3D convolutional layer. To demonstrate the effectiveness, experiments have been conducted on two popular public benchmarks, CASIA-B and OU-MVLP, and our approach is demonstrated superior performance, comparing with the state-of-art methods.

[*] Corresponding author. E-mail address: heukejun@126.com

# 1. Introduction

As a promising biometric character, gait is a kind of unique approach to identification. This is because video-based recognition just demands a camera, and the identification of the person could be known from a long distance[1]. Gait recognition does not require the object's explicit cooperation, allowing it to be applied widely in access controlling, person identification, and crime prevention.

The essence of the video-based gait recognition is to exploit the behavioral pattern of walking which is a kind of distinctive evidence to indicate the identification[2]. It is such motion learning that enables gait to be distinguished from other appearance-based biometric, like faces, fingerprints, and iris, thus being given an exclusive edge that the work can be done with low-resolution images[3]. Therefore, the key to gait recognition is to learn the discriminative behavioral features.

However, understanding a sequence of silhouettes and then distilling the behavioral representation is a great challenge [4]. Some previous works mostly focus on extracting the robust appearance feature[5],[6],[7]. In this case, the gait video is compressed as one image[8],[9], like gait energy image (GEI)[10], or regarded as a set of shuffled images, followed by a special and elaborate convolutional neural network (CNN) to identify[4],[11]. Additionally, some other motion-based methods most adopt the body skeleton information[12]. These methods show more robustness to variations, such as cloth and carrying changing[13],[14], but relying on pose estimation accuracy and completely ignoring fine-grained spatial information. Also, there are methods employing 3D-CNN [15]and LSTM[16] but they are difficult to train and bring a higher computational cost.

To alleviate these issues, it is imperative to explore a simple and effective approach to learn spatiotemporal information. In this paper, we propose a novel sequential convolutional network (SCN) where the classic CNN is optimized to comprehend the motion feature in time series. To be specific, spatiotemporal information learning is carried out by two sub-blocks: behavioral information extractor (BIE) and multi-frame aggregator (MFA). As shown in Fig. 1, a weight-sharing CNN is applied to extract spatial features on each image. Then BIE is capable of analyzing the relationship between frames in the time series by learning from a behavior template that represents the motion. Lastly, MFA would integrate and distill all features to identify. In general, the proposed SCN offers three contributions::

- In BIE, three types of templates are proposed to represent the short-range or long-range behavior features, and three kinds of structures are designed to fuse temporal features and spatial features. In this case, BIE, a basic convolutional module, can understand and process motion information on a gait sequence, making it available for the convolutional backbone without modification.
- In MFA, the mobile aggregator based on the 3D convolution method is proposed to deal with the variable-length gait video, allocating weights for frames according to the importance of information and the quality of the image, thereby synthetic features to represent the entire sequence.
- Rigorous ablation experiments conducted on CASIA-B further verifies the effectiveness of each part in SCN. Besides, extensive experiments on CASIA-B[17] and OU-MVLP[18] exemplify SCN outperforms other state-of-the-art methods.

# 2. Related work

In this section, an overview of deep learning gait recognition methods is presented. The existing approaches can be classified into two categories: appearance-based gait recognition and motion-based recognition.

2.1 Appearance-based gait recognition

There are two kinds of appearance-based gait recognition methods. The difference exists in the form of the input: an image, i.e. a template where the video frames are aggregated, or an original sequence.

There are a wide variety of classic templates, and the image is calculated on the pixel level, such as GEI[10], active energy image (AEI)[19], and chrono-gait image (CGI)[9]. Next, a network is designed to learn the gait template. Shirage et al.[20] directly applied CNN to extract gait features from GEIs, namely GEInet. Wu et al.[21] designed a CNN network for gait recognition, and comprehensive experiments showed excellent cross-view performance. Yu et al.[22] applied stacked progressive auto-encoders to convert the gait features from one view to another to reduce the effect of view changing. Song et al.[23] proposed the joint learning to integrate contour segmentation and gait recognition in an end-to-end network. Takemura et al.[24] analyzed the tradeoff led by the subject difference and the view difference (2dif), using a triplet of inputs and triplet ranking loss for identification (3in). Generative adversarial network (GAN) was also adopted in gait recognition to learn the invariant feature for identification. Discriminant gait GANs (digGAN)[25], two-Stream GANs[26], and multi-task GANs[27] were respectively employed to enhance the robustness under cross-condition scenarios. Additionally, the original sequence can be considered as a set of images, which could reduce the loss of information in obtaining templates[4]. Chao et al.[28] proposed a network named GaitSet to extract identity information from the set, achieving satisfactory results, but the relationship in time series was neglected. Sepas-Moghaddam et al.[29] applied a bidirectional recurrent neural network to learn the relations between partial features distilled in a sequence (PartialRNN). Appearance-based methods could extract the information in each silhouette, which is of paramount importance for gait recognition.

2.2 Motion-based recognition

There are also two groups of motion-based recognition methods: the first is to employ human pose to learn the gait motion; the second is to apply deep networks that can have the memory with temporal information.

Due to pose estimation approaches [12, 30], the body structure can be easy to attain, and it contributes to learning the motion of people, reducing the disturbance of the covariate. An et al. [31] presented a large-scale human pose-based gait database and a simple CNN (CNN-Pose) to utilize the pose information. Li et al. [13] proposed a skinned multi-person linear to learn pose and shape features, and a pre-trained human mesh recovery could estimate parameters. Liao et al.[14] exploited human 3D pose and RNN or LSTM to learn spatiotemporal gait features, i.e., PoseGait. Besides, Wu et al.[21] also directly attempted to applied 3D CNN for gait recognition. Zhang et al.[16, 32] designed an auto-encoder framework to disentangle pose and appearance features, and integrated LSTM-based pose features to form the gait representation. Wang et al.[33] proposed event-based gait recognition (EV-Gait) method to learn the motion consistency, using dynamic vision sensors. Motion information can reflect the behavioral feature of objects' identity, and it is imperative to explore a simple and effective means to make full use of hidden motion features in a sequence.

# 3 Method
## 3.1 Overview

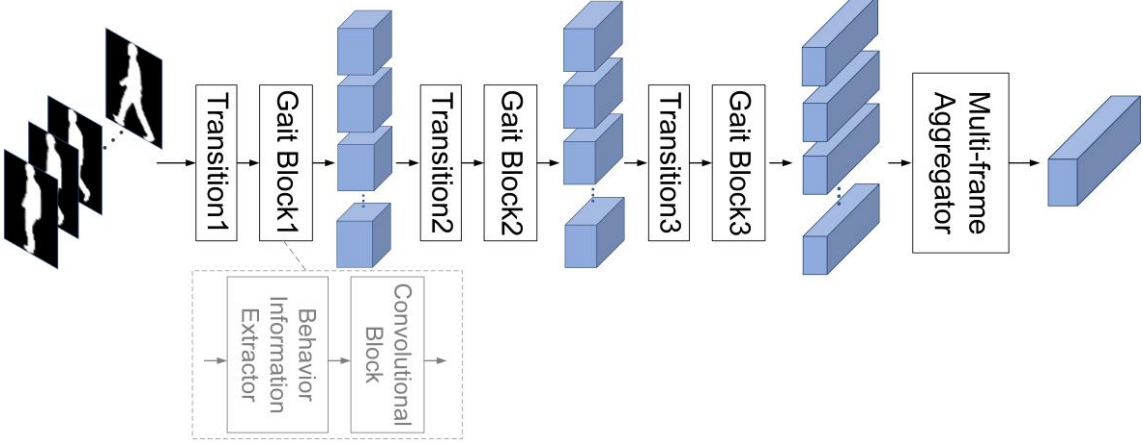

Fig. 1. The pipeline of the proposed method.

Fig. 1 illustrates the specialized network structure, the sequential convolutional network (SCN) for gait recognition. To be specific, the input is a sequence of silhouettes, and then each frame is fed into three groups of analogous structures in order. Each structure contains a transition and a gait block (TGB), hence extracting temporal and spatial features of each frame. Finally, we construct a multi-frame aggregator (MFA) to integrate all features in a sequence. In this case, the process can be formulated as:

$$feature_{output} = SCN(S) = MFA(TGB(S)) = MFA(tgb_3(tgb_2(tgb_1(S)))) \quad (1)$$

where $feature_{output}$ denotes the final extracted features of the input sequence, $S = \{f_i \mid i = 1, 2, 3, \cdots, n\}$, including $n$ frames. Features of every frame $f_i$ can be fused by MFA. TGBs can distill the spatiotemporal representation. TGBs are made up of sub-blocks tg*b,* and the *tgb* consists of the behavioral information extractor (BIE) and the convolutional block (CB):

$$F = TGB(S) = tgb_3(tgb_2(tgb_1(S))) \quad (2)$$

$$tgb(F_{input}) = CB(BIE(F_{input})) \quad (3)$$

where $F$ denotes a set of features of all frames $F = \{F_i \mid i = 1, 2, 3, \cdots, n\}$. For the input $F_{input}$ of *tgb*, it would be calculated by the function of BIE to pay more attention to the motion information, whose specific method would be described in section 3.2.

Furthermore, the function MFA is employed to integrate $F$ to yield the sequence-level features in three dimensions, and its details would be shown in section 3.3.

For gait recognition, the dataset is divided into a gallery set and a probe set, where each sequence $S_j$ corresponds to an identity $y_j$ ($j \in 1, 2, \ldots, N$). Therefore, the purpose of the above entire process can be regarded as:

$$\hat{y} = \arg\max_{y_j} P(y_j \mid F_{output, probe_j}, F_{output, gallery}) \quad (4)$$

where $P(y_i \mid F_{output, probe_i}, F_{output, gallery})$ is the probability of final gained features of sequence $probe_j$, $F_{output, probe_j}$, belonging to identity $y_j$ in features of the gallery, $F_{output, gallery}$.

3.2 Behavioral information extractor based on motion templates

Since classical convolutional networks suffer difficulty in processing continuous video information, it is more suitable for the task of the interpretation of a single image, rather than understanding the correlation between frames. Accordingly, the convolutional network focuses more on the appearance in each image to identify people. In fact, gait recognition is expected to utilize habits and behaviors for identification when people are walking, so the BIE, a novel method of convolution, is proposed in this paper. It first calculates motion templates through a set of serial features in a sequence, then enables the motion information represented by templates to be introduced to original features, thereby making it available for convolution to learn temporal information. In this method, we design three types of motion templates and three kinds of introducing approaches, and the comparison will be shown in section 4.3.

3.2.1 Motion template
Motion templates aim to extract motion features from a sequence of features. Because convolutional networks could not understand the order and the relation between frames, templates ($T$) are employed to explore and analyze the relevance of features of images. Thus, this paper constructs three types of templates to express such relevance, based on difference, multi-difference, and removing static information, respectively.
(1) Template based on the difference of features between adjacent frames

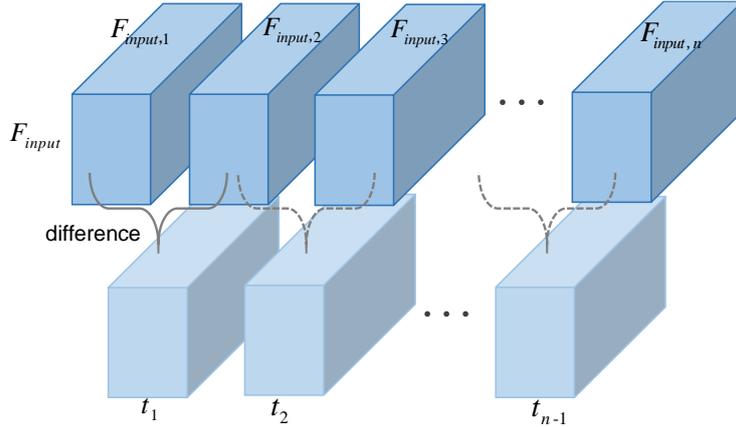

Fig. 2 Template based on the difference of features between adjacent frames

The feature map extracted from each frame represents the abstract information of each gait contour. Because the features are learned by the weight-sharing network, the difference between adjacent features could reflect the changing, i.e. motion information. According to this method, the difference-based template can be obtained as shown in Fig. 2, and the $k$th template $t_{d,k}$ formulated as:

$$t_{d,k} = \left| F_{input,k+1} - F_{input,k} \right| \tag{5}$$

$$T_d = \{t_k \mid k = 1,2,3,\ldots,n-1\} \tag{6}$$

where $F_{input,k}$ is the feature map of the kth frame, and $T_d$ denotes the difference-based template. It is worth noting that, for $n$ as the length of the sequence, the length of $T_d$ corresponds to $n$-1.
(2) Template based on multi-difference of features between adjacent frames

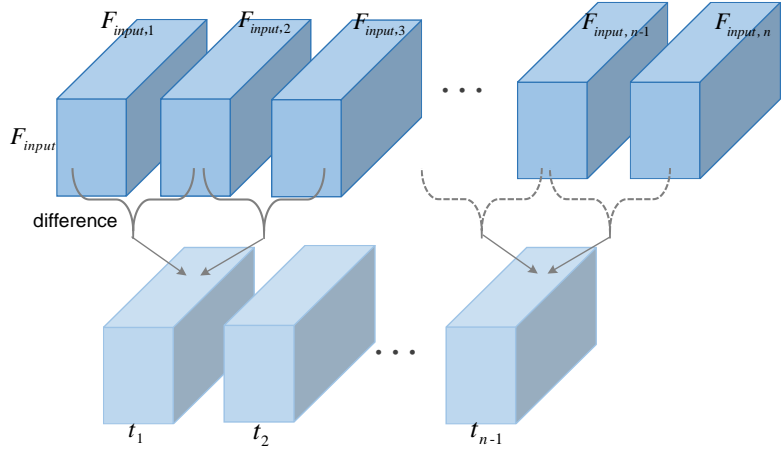

Fig. 3 Template based on multi-difference of features between adjacent frames

Since regarding two frames as a motion unit might be too small, we employ the multi difference to cover more motion information. As shown in Fig. 3, the two differences are accumulated. Similarly, $t_{md,k}$, $T_{md}$ can be formulated as:

$$t_{md,k} = \left|F_{input,k+1} - F_{input,k}\right| + \left|F_{input,k} - F_{input,k-1}\right| \qquad (7)$$

$$T_{md} = \{t_{md,k} \mid k = 2, 3, 4, \ldots, n-1\} \qquad (8)$$

where the first one in $T_{md}$, $t_{md,1}$, is calculated based on the 2nd frame, and the length of $T$ corresponds to $n$-2.

(3) Template based on static features exclusion

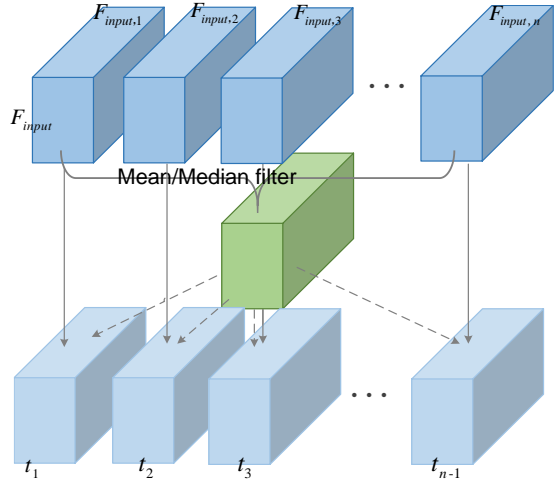

Fig. 4 Template based on static features exclusion

Actually, the input includes both static features and dynamic features, but they are highly coupled. If they can be decoupled, the motion information can be emphasized and learned better. Given that the static features of an individual are identical and universal in each frame of a sequence, the common sharing features can be regarded as the approximation of static information. As illustrated in Fig. 4, the mean or median filter is deployed to extract static features. Then, the difference between the original features of each frame and the static features can express the dynamic information, $T_{se}$, formulated as:

$$t_{se,k} = \left|F_{input,k} - median(F_{input})\right| \qquad (9)$$

$$\text{or} \quad t_{se,k} = |F_{input,k} - mean(F_{input})| \tag{10}$$

$$T_{se} = \{t_{se} \mid k = 1, 2, 3, \ldots, n\} \tag{11}$$

where mean(•) and median(•) denotes the mean function and the median function separately. Moreover, by this means, the information of the entire sequence is fused into every template, i.e., long-range information is introduced into features of every frame.

3.2.2 Introducing approach

After gaining the template, how to introduce and fuse the template into the network, and make full use of it, is the following difficulty. In this paper, we present three kinds of approaches: micro motion fusion, global motion fusion, and adaptive fusion.

(1) Micro motion fusion:

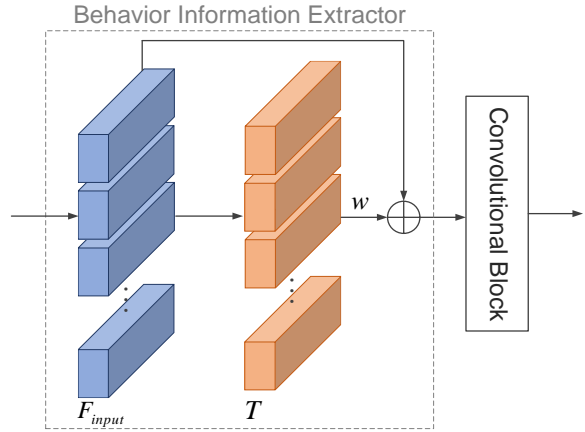

Fig. 5 Micro motion fusion

Micro motion fusion is an introducing approach where the features of $k$th frame $F_{input,k}$ is only associated with its corresponding templates $t_k$, with no relation to other frames' motion templates. As shown in Fig. 5, this process can be formulated as:

$$BIE(F_{input,k}) = F_{input,k} + w \times t_k \tag{12}$$

where $w$ is a trainable weight. During this process, the length of $F_{input}$ is supposed to be truncated to correspond with the length of $T$ in the addition operation.

(2) Global motion fusion:

The gait sequence is represented by contour images, but the accuracy of the contour is influenced by the segmentation method. Thus, it is necessary to consider the impact with it. Especially for short-range motion templates adopted, the motion features accompany with the disturbance of segmentation, as we could not estimate that the change is caused by the action or the segmentation error. Under this circumstance, micro motion fusion is sensitive to such noise and might not be robust, so the global motion fusion is proposed, as illustrated in Fig. 6, formulated as:

$$BIE(F_{input,k}) = F_{input,k} + w \times 1\_1conv(cat(mean(T), \max(T))) \tag{13}$$

where $T$ is calculated by the mean(•) and the max(•) in the frame dimension, so the global information of $T$ is extracted and introduced into $F_{input}$. Furthermore, another advantage of this approach to overcome the absence of inner information in the contour representation. This is because the change of contours could reflect the motion of the outer profile, but in fact, there is the inner action neglected, since it is always shown as the inner identical pixel value between adjacent frames. Therefore, this approach allows more hide information to be involved from the extensive and global perspective.

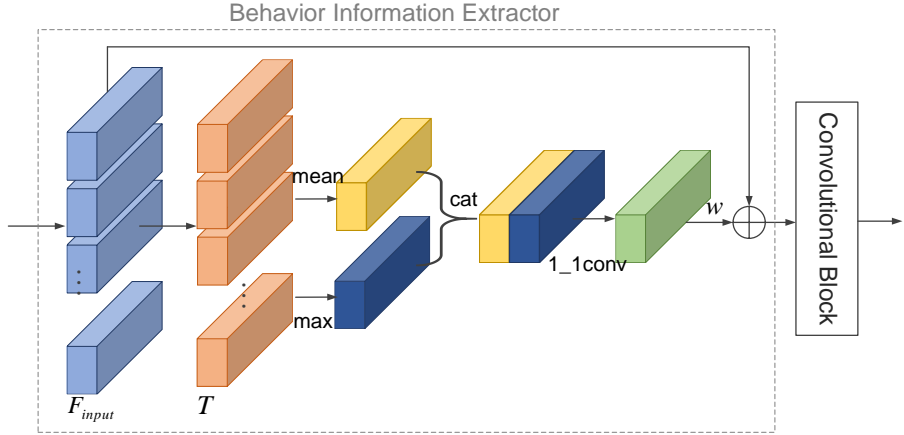

Fig. 6 Global motion fusion. Cat indicates concatenate, and 1_1conv means the convolutional layer with the 1×1 kernel.

 (3) Adaptive fusion:

The micro motion fusion might be likely to be disturbed by the inaccurate segmentation, while the global motion fusion might give the same motion features to every frame, ignoring the sequential information in a short range. Thus, we propose the adaptive fusion approach. As shown in Fig. 7, the template $T$ is fed into the convolutional block ($CB_T$) which has the same structure as the backbone, so this process can be formulated as:

$$CB(BIE(F_{input,k})) = CB(F_{input,k}) + CB_T(t_k) \tag{14}$$

where $CB_T$ could learn suitable weights to further represent the motion features in $T$, because some noise in the template could be filtered by $CB_T$. Then the motion features and the frame-appearance features are combined.

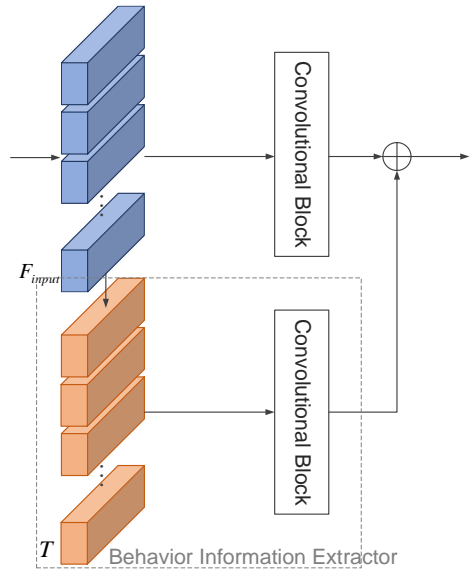

Fig. 7 Adaptive fusion

3.3 Multi-frame aggregator

Gait recognition is essentially a video understanding task. The above steps can merely learn the feature map of every frame, although some related inter-frame information is considered. The purpose of MFA is to integrate all frame-level information $F$ to learn the discriminative sequence-level features $feature_{output}$:

$$feature_{output} = MFA(F) \tag{15}$$

where $F = \{F_i \mid i = 1, 2, 3, \cdots, n\}$. Due to the changing distance between the pedestrian and the camera, gait silhouettes vary in size. So the silhouettes are usually aligned into the same size, which leads to different image qualities. (Some images are enlarged by bilinear interpolation, ending up with low qualities, whereas others are unchanged or compressed, containing more accurate information.) Hence, MFA is expected to evaluate $F_i$, and fuse them in a feature-wise means. Besides, since the length of the gait video, $n$, is uncertain in real life (the number of a person's gait contours can be arbitrary), it is difficult to deploy a fixed map as MFA. In recent work, some classic approaches directly apply statistical functions[11, 28, 29], like max(•) and mean(•), but such approaches are too straightforward to fuse the frame-level information. In this case, we design a mobile aggregator to address this problem.

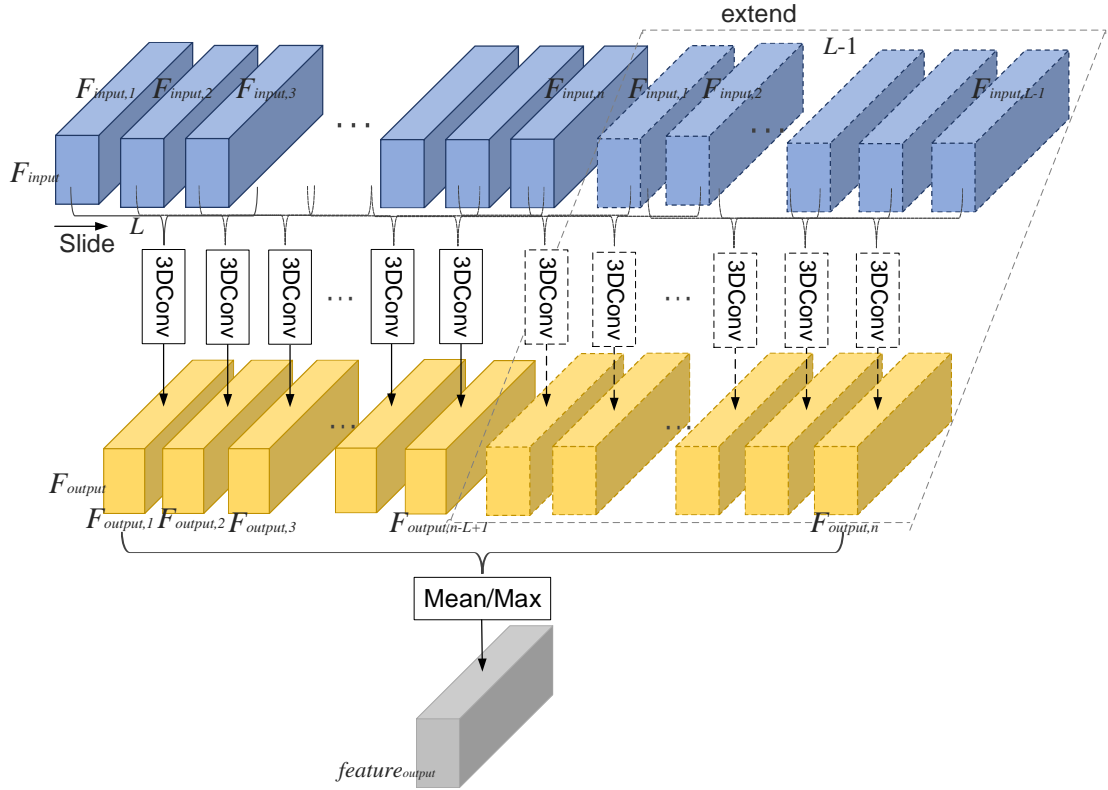

Fig. 8 Multi-frame aggregator

As shown in Fig. 8, starting with the first feature, $F$ are cut out into segments of length $L$, and then the segment is fed into a 3D convolutional layer, yielding a corresponding integrated feature, $F_{output,1}$. After that, the 3DConv slides with stride 1, and the next segment is fed. Concretely, the $j$th segment $F_{S,j} = \{F_j \mid j = j, j+1, j+2, \cdots, j+L-1\}$ ($j=1,2,3,\ldots,n-L+1$) is sent to the 3DConv orderly. That is, the 3DConv is sliding on $F$ and the length of the sliding window is $L$, so $F_{output,j}$ is learned from each segment $F_{S,j}$. However, in this case, the maximum of $j$ is $n-L+1$, rather than $n$, ending up that the size of $F_{output}$ is $n-L+1$. During this process, the middle features in $F$ are fed and calculated more times, while the feature in either end is less involved in the calculation, hence the imbalance on some frames. Thus, the $F$ is extended as illustrated in Fig. 8. To be specific, *F* becomes a cycle, where features from the 1st to the *L*-1th are concatenated at the end of *F*. Accordingly, the 3DConv is carried out *n* times, and the shape of results is the same as *F*. In this way, every frame-level feature is given fair weights. Finally, max(•) or mean(•) is adopted on sequence dimension on $F_{output}$, thereby the discriminative sequence-level features $feature_{output}$.

## 4 Experiments

The proposed method is evaluated on CASIA-B dataset[17] and OU-MVLP[18] dataset. Firstly, comprehensive ablation studies are conducted on CASIA-B to prove the effectiveness of each component (BIE and MFA) in SCN. Under this circumstance, results of templates, introduction approaches, and MFA in terms of different settings are illustrated and analyzed. Then, the performance of SCN is compared with state-of-the-art gait recognition methods on CASIA-B and OU-MVLP.

### 4.1 Datasets

CASIA-B dataset is one of the most popular gait dataset, consisting of 124 subjects where each subject contains 11 (0°, 18°, 36°,…, 180°) walking views and 3 walking conditions (normal case, carrying bag, cloth changing). And each subject under each view has 6 sequences in the normal case (NM), 2 sequences on carrying bags (BG), and 2 sequences on cloth changing (CL) (wearing coats or jackets), so one subject has 10 sequences on one certain view, and altogether 110 sequences (10×11=110). A test protocol that has widely been adopted is applied in this paper[21],[28],[11]. In this protocol, the first 74 subjects are deployed for the train set, and the rest 50 subjects are reserved for testing. During testing, the first 4 sequences of NM (NM 1-4) form the gallery set, whereas the rest 6 sequences are included in the probe set, divided into three sub-sets: the NM sub-set containing NM 5- 6, the BG sub-set containing BG 1-2, and the CL sub-set containing CL 1-2.

OU-MVLP dataset [18] is currently the largest public gait dataset, consisting of 10307 subjects, 14 different angles, ranging from 0° to 90° (0°,15°,…,90°), and 180° to 270° (180°,195°,…,270°), and it involves 2 sequences (00 and 01) per view. A prevalent test protocol is applied in this paper[20], [24], where the first 5153 subjects are deployed for the train set, and the rest 5154 subjects are reserved for testing and four viewing angles, 0°, 30°, 60°, and 90°, have been deployed for cross-view gait recognition. During testing, the first sequence of each subject (00) forms the gallery set, while the rest sequences (01) are regarded as the probe set.

### 4.2 Training Details

#### 4.2.1 Network.

The structure of the transition and the convolutional block in SCN is based on residual learning, as shown in Fig. 9, made up of convolutional layers, Max Pooling Layers, and Leaky ReLU activation function. (In the first transition, there is not the Pool layer.) The output's channel of the three blocks is 32, 64, and128 in order. (For OU-MVLP, because data are 20 times than CASIA-B, the output's channel of the three blocks is 64, 128, and 256 in order.) Besides, the kernel of the 3D convolutional layer is 3×1×1 with stride1, and the output shape is equal to the input shape.

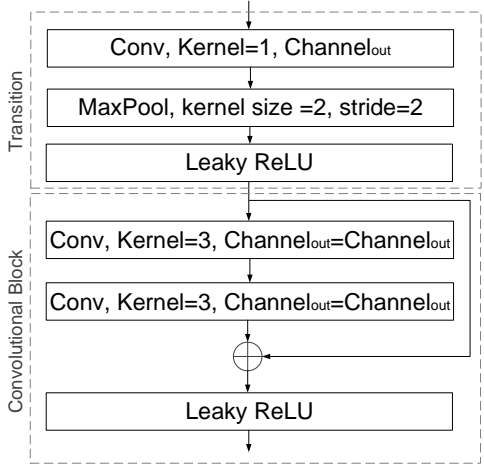

Fig. 9 Convolutional Block

4.2.2 Loss.

Triple loss is very suitable for tasks where samples have small differences[34]. Three types of samples are fed, the anchor sample, the positive sample, and the negative sample with a different identity of the anchor. It can be defined as:

$$L_{triplet} = \frac{Max(M - L2_{anch,pos} + L2_{anch,neg}, 0)}{2M} \quad (16)$$

where a margin $M$ is to enable $L2$ distance between features belonging to a subject and other features for the same subject smaller than that between features belonging to a subject and other features for a different subject.

4.2.3 Implementation Details.

First, gait bounding boxes of input silhouettes are obtained, and then silhouette frames should be aligned and resized as 64×44. The length of the segment in training is set as 30 (30 serial frames in a sequence selected randomly). Concretely, the raw sequences whose lengths are less than 15 frames should be discarded, and ones whose lengths are more than 15 frames but less than 30 frames should be sampled repeatedly. $M$ in the triplet loss is set as 0.2. The optimizer is Adam with the momentum of 0.9. Batch All (BA$_+$) triplet loss is adopted in training[35], where the batch size is $p \times k$, and $p$ denotes the number of subjects and $k$ denotes the number of segments of each subject in the batch. The batch size, learning rate, and the number of iterations are different for the two datasets, due to the difference of the amount of data. For CASIA-B, the batch size is (8,6) because of the limit of the hardware, the number of iterations is 150K, and the learning rate first is 1e-3, reduced to 1e-4 after the 10Kth iteration, and lastly reduces as 1e-5 after the 80Kth iteration. For OU-MVLP, the batch size is (6,4) owing to the limit of hardware, the number of iterations is 300K, and the learning rate first is 1e-3, reduced to 1e-4 after the 50Kth iteration, and lastly reduces as 1e-5 after the 200Kth iteration.

At the test phase, the aligned and resized gait sequence can be fed into the model directly, and then the distance between gallery and probe is calculated by the Euclidean distance of the corresponding feature maps. Lastly, the Rank 1 recognition accuracy can be yielded.

**4.3 Ablation Study**

To verify the effectiveness of each innovation in section 3, we have conducted several ablation experiments with various settings on CASIA-B.

4.3.1 Impact of BIE
Table 1.
Ablation experiments conducted on CASIA-B in terms of various setting on BIE. T1, T2, and T3 denote three types of proposed templates mentioned in section 3.2: the template based on difference of features between adjacent frames, the template based on multi-difference of features between adjacent frames, and the template based on static features exclusion.

| Template | | | BIE Introduction approach | | | NM | BG | CL |
|---|---|---|---|---|---|---|---|---|
| T1 | T2 | T3 | Micro fusion | Global fusion | Adaptive fusion | | | |
| | | | | | | 92.2 | 83.5 | 65.0 |
| √ | | | √ | | | 92.0 | 83.5 | 65.5 |
| | √ | | √ | | | 92.2 | 83.4 | 65.7 |
| | | √ | √ | | | 92.5 | 84.1 | 66.8 |
| √ | | | | √ | | 92.1 | 84.0 | 66.7 |
| | √ | | | √ | | 92.1 | 84.1 | 66.5 |
| | | √ | | √ | | 92.4 | 84.8 | 67.8 |
| √ | | | | | √ | 94.0 | 87.1 | 69.2 |
| | √ | | | | √ | **94.2** | 87.3 | 69.0 |
| | | √ | | | √ | 94.1 | **87.4** | **69.5** |

Table 1 shows 9 results on different settings according to three types of templates and three kinds of introducing approaches proposed in Section 3.2. The result is the averaged rank-1 accuracy on all 11 views except identical-view cases. To begin with, the results on the original network without BIEs are presented in the first line. As comparison, introducing BIEs outperforms the original, enhancing 2%, 4%, and 5% in the best case on NM sub-set, on BG sub-set, on CL sub-set, respectively. It can be concluded that BIE is conducive to distill the spatiotemporal information to boost performance. BIE could fuse the behavioral information, which is of top priority for gait recognition. Moreover, for different templates, T3 yields better results no matter which introducing approach is selected. This might be because T3 is calculated based on the whole sequence, where the long-range motion information is covered, while T1 and T2 pay more attention to the local motion of gait. Additionally, focusing on the entire sequence allows T3 robust, unlike that T1 and T2 are distracted by the disturbance of body segmentation. Lastly, for the introducing approach, adaptive fusion is the best, achieving great improvement, compared with the other two approaches. This could be caused by that convolution block can further extract the effective features, reducing the noise in the template, whereas the micro fusion simply adds the template and the global fusion, and global fusion neglects the local relationship between frames in a short range.

4.3.2 Impact of MFA
In Table 2, the results of adding MFA are shown. Firstly, the first line and the second line compare different statistical functions (mean(•) and max(•)), and it can be seen that the network employing max(•) performs better to aggregate gait feature maps on multi frames in a sequence, which is consistent with [11] [28]. In addition, MFA can yield better performance by comparing the first two lines and the second two lines. Especially for setting with mean(•), the accuracy is separately increased almost 1.5%, 4.5%, and 6% on three sub-sets. This might be because that the important frame that contains discriminative features is assigned a larger weight by the mobile 3D convolutional layer, thereby enhancement of the key frame to achieve better results. Finally, BIEs are added in the last two lines, illustrating the final results of the proposed method. (T3 and the adaptive fusion are applied in the BIE, according to the above experiment.) It can be observed that combining the MFA and the BIE yield the best result, achieving 95.2%, 89.8%, 73.9% on three sub-sets respectively.

Table 2. Ablation experiments conducted on CASIA-B in terms of various settings on MFA.

| Statistical function | | BIE | MFA | NM | BG | CL |
|---|---|---|---|---|---|---|
| Mean | Max | | | | | |
| √ | | | | 92.2 | 83.5 | 65.0 |
| | √ | | | 94.0 | 88.0 | 71.7 |
| √ | | | √ | 93.8 | 87.9 | 71.0 |
| | √ | | √ | 94.2 | 88.3 | 72.1 |
| √ | | √ | √ | 94.3 | 88.8 | 72.1 |
| | √ | √ | √ | **95.2** | **89.8** | **73.9** |

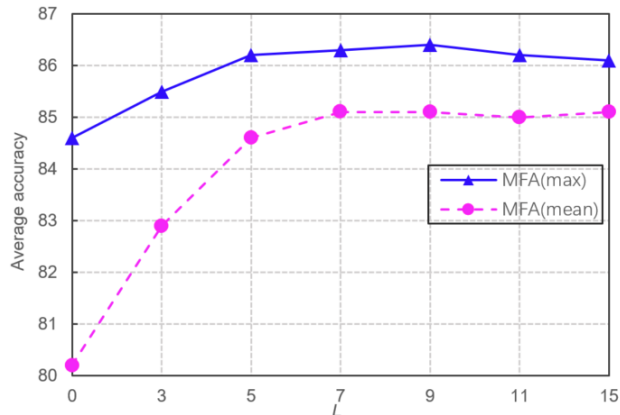

Fig. 10 Impact of $L$ in MFA

The length $L$ of segments of $F$ (the length of the sliding window) is responsible for the performance of the MFA. Fig. 10 presents the average accuracy of three sub-sets in terms of different setting of $L$. It can be observed that the recognition accuracy grows with the increase of $L$, but the accuracy reaches stability when $L$ is big enough. Through experiments, 7 is an optimal value for $L$ considering the computational cost and performance. (The $L$ is set to 7 in experiments shown in Table 2.) Furthermore, the setting value of $L$ could be closely related to the dataset. In CASIA-B, 7 frames include more than a process where one leg takes a step, so this setting can achieve a better performance in this case. Besides, BIE blocks employing T3 to add long-range information could allow the MFA to cover frames in a relatively short range.

### 4.4 Comparison with Other Methods

Table 3, Table 4, and Table 5 compare the performance between the state-of-the-art methods and SCN on the NM subset, the BG subset, and the CL subset. Results are also the average on the 11 gallery views except for the identical view (the view probe is equal to the gallery view).

It can be seen that our method meets a new state-of-the-art in Table 3, as same as PartialRNN, but our method is better in more views. So the result can demonstrate the effectiveness of the proposed SCN, and the spatiotemporal feature can be learned. For CNN-LB and Gait-joint, their inputs are gait images that can represent a video, GEI, rather than a sequence. Such pre-process can reduce the complexity and the computational cost, but it may ignore motion features, leading to less unideal accuracy than our method. GaitSet and PartialRNN regard a gait sequence as a set of shuffled images, and the partial method is applied in PartialRNN to enhance the robustness. Although they can extract silhouette features by the elaborate network, especially PartialRNN yields results as excellent as SCN, it might neglect the relationship between frames which indicates the behavioral pattern of the walking. It is worth noting that PartialRNN and SCN address the hinder in gait recognition from different perspectives. SNC does not pay too much attention to appearance. Instead, it can analyze the order in the time series to learn the behavioral information. PoseGait employs the skeleton to represent the motion of objects, which is simple and efficient, but the ignored body appearance lowers accuracy, inferior to our method which combines the motion information and the appearance information. CNN-3D and GaitNet can learn the motion feature by 3D convolution and LSTM, which is complicated and might give rise to a calculation burden. Our method just bases on convolutional layers, which is easy to implement.

Table 3.
Comparison with Other Methods on CASIA-B on NM sub-set

| Method | View | | | | | | | | | | | mean |
|---|---|---|---|---|---|---|---|---|---|---|---|---|
| | 0° | 18° | 36° | 54° | 72° | 90° | 108° | 126° | 144° | 162° | 180° | |
| CNN-3D[21] | 87.1 | 93.2 | 97.0 | 94.6 | 90.2 | 88.3 | 91.1 | 93.8 | 96.5 | 96.0 | 85.7 | 92.1 |
| CNN-LB[21] | 82.6 | 90.3 | 96.1 | 94.3 | 90.1 | 87.4 | 89.9 | 94.0 | 94.7 | 91.3 | 78.5 | 89.9 |
| EV-Gait[33] | 77.3 | 89.3 | 94.0 | 91.8 | 92.3 | 96.2 | 91.8 | 91.8 | 91.4 | 87.8 | 85.7 | 89.9 |
| Gait-joint[23] | 75.6 | 91.3 | 91.2 | 92.9 | 92.5 | 91.0 | 91.8 | 93.8 | 92.9 | 94.1 | 81.9 | 89.9 |
| GaitSet[28] | 90.8 | 97.9 | 99.4 | 96.9 | 93.6 | 91.7 | 95.0 | 97.8 | **98.9** | 96.8 | 95.8 | 95.0 |
| GaitNet[32] | **93.1** | 92.6 | 90.8 | 92.4 | 87.6 | **95.1** | 94.2 | 95.8 | 92.6 | 90.4 | 90.2 | 92.3 |
| PartialRNN[29] | 91.1 | 98.0 | 99.4 | **98.2** | 93.2 | 91.9 | **95.2** | 98.3 | 98.4 | 95.7 | 87.5 | **95.2** |
| PoseGait[14] | 55.3 | 6936 | 73.9 | 75.0 | 68.0 | 68.2 | 71.1 | 72.9 | 76.1 | 70.4 | 55.4 | 68.7 |
| SCN(Ours) | 89.7 | **98.5** | **99.8** | 97.9 | **94.4** | 91.2 | 94.5 | 97.1 | 97.6 | **97.0** | 89.4 | **95.2** |

Table 4.
Comparison with Other Methods on CASIA-B on BG sub-set

| Method | View | | | | | | | | | | | mean |
|---|---|---|---|---|---|---|---|---|---|---|---|---|
| | 0° | 18° | 36° | 54° | 72° | 90° | 108° | 126° | 144° | 162° | 180° | |
| CNN-LB[21] | 64.2 | 80.6 | 82.7 | 76.9 | 64.8 | 63.1 | 68.0 | 76.9 | 82.2 | 75.4 | 61.3 | 72.4 |
| GaitSet[28] | 83.8 | 91.2 | 91.8 | 88.8 | 83.3 | 81.0 | 84.1 | 90.0 | 92.2 | 94.4 | 79.0 | 87.2 |
| GaitNet[32] | 83.0 | 87.8 | 88.3 | 93.3 | 82.6 | 74.8 | 89.5 | 91.0 | 86.1 | 81.2 | 85.6 | 85.7 |
| PoseGait[14] | 35.3 | 47.2 | 52.4 | 46.9 | 45.5 | 43.9 | 46.1 | 48.1 | 49.4 | 43.6 | 31.1 | 44.5 |
| PartialRNN[29] | 86.0 | 93.3 | 95.1 | 92.1 | **88.0** | **82.3** | **87.0** | 94.2 | 95.9 | 90.7 | 82.4 | 89.7 |
| SCN(Ours) | **86.7** | **94.6** | **96.0** | **92.5** | 85.8 | 80.5 | 84.9 | 91.5 | **96.0** | **93.1** | **86.0** | **89.8** |

In Table 4, our method achieves 89.8%, outperforming other state-of-the-art methods, and the superiority of our model is evident on BG sub-set. The reason might be that appearance is changing when people carry a bag. Under this circumstance, focusing on silhouettes may end up a weak adaptation, whereas the capability of SCN that can draw on motion information could make up for this weakness. It is noteworthy that PartialRNN yields better results under side angles (near 90°), while our method is better on other views. This might be because silhouettes change so much on side angles that SCN is distracted, since the learning of behavior is based on analysis of silhouettes. However, by partition, PartialRNN can decrease the disturbance of the changing. Similarly, this may be the reason why it achieves the best result under CL condition, as shown in Table 5. Nonetheless, it is worth noting that the partial method does not have conflicts with our SCN. By introducing this partial approach, SCN might achieve better results on CL sub-set.

Table 5.
Comparison with Other Methods on CASIA-B on CL sub-set

| Method | View | | | | | | | | | | | mean |
|---|---|---|---|---|---|---|---|---|---|---|---|---|
| | 0° | 18° | 36° | 54° | 72° | 90° | 108° | 126° | 144° | 162° | 180° | |
| CNN-LB[21] | 37.7 | 57.2 | 66.6 | 61.1 | 55.2 | 54.6 | 55.2 | 59.1 | 58.9 | 48.8 | 39.4 | 54.0 |
| GaitSet[28] | 61.4 | 75.4 | 80.7 | 77.3 | 72.0 | 70.1 | 71.5 | 73.5 | 73.5 | 68.4 | 50.0 | 70.4 |
| GaitNet[32] | 42.1 | 58.2 | 65.1 | 70.7 | 68.0 | 70.6 | 65.3 | 69.4 | 51.5 | 50.1 | 36.6 | 58.9 |
| PoseGait[14] | 24.3 | 29.7 | 41.3 | 38.8 | 38.2 | 38.5 | 41.6 | 44.9 | 42.2 | 33.4 | 22.5 | 36.0 |
| PartialRNN[29] | **65.8** | **80.7** | **82.5** | **81.1** | **72.7** | 71.5 | 74.3 | 74.6 | 78.7 | 75.8 | 64.4 | **74.7** |
| SCN(Ours) | 63.7 | 79.2 | 82.3 | 77.7 | 69.4 | 71.5 | 73.5 | **77.9** | 78.4 | **76.5** | 62.4 | 73.9 |

In order to evaluate the generalization of the proposed method, the experiment of SCN is conducted on OU-MVLP which is the largest public gait dataset [18], containing the largest number of objects. Table 6 presents the comparison. To keep a fair comparison, the protocol mentioned is strictly followed on cross-view condition, i.e. four typical views are adopted for the gallery set, 0°, 30°, 60°, 90°. Table 6 shows SCN obtains an outstanding result, where the accuracy exceeds 77% on all four views, suggesting that our method is able to generalize well on the dataset with such a large scale.

Table 6. Comparison with Other Methods on OU-MVLP

| Method | View | | | | mean |
|---|---|---|---|---|---|
| | 0° | 36° | 60° | 90° | |
| GEINET[20] | 8.2 | 32.3 | 33.6 | 28.5 | 25.7 |
| CNN-LB[21] | 14.2 | 32.7 | 32.3 | 34.6 | 28.5 |
| 3in+2diff[24] | 25.5 | 50.0 | 45.3 | 40.6 | 40.4 |
| DigGAN[25] | 30.8 | 43.6 | 41.3 | 42.5 | 39.6 |
| GaitSet[28] | 77.7 | 86.9 | 85.3 | **83.5** | 83.4 |
| CNN-Pose[31] | 47.3 | 69.1 | 73.2 | 49.0 | 59.7 |
| SCN(Ours) | **78.6** | **87.4** | **85.9** | 83.2 | **83.8** |

## 5 Conclusion

In this paper, we present a novel perspective on gait recognition that a sequential convolutional network (SCN) can be constructed to extract spatiotemporal information based on the convolutional backbone, ending up learning the walking pattern of individuals. The proposed method contains behavioral information extractors and a multi-frame aggregator. Behavioral information extractors can comprehend intermediate feature maps in time series by motion templates where the relationship between frames is analyzed, hence distilling the information of walking patterns. The multi-frame aggregator can integrate features on a variable-length sequence by a mobile 3D convolutional layer. In final, experiments conducted on two prevalent databases, CASIA-B and OUMV-LP, comprehensively verify the superior performance of SCN as well as its components.